\documentclass[12pt]{article}
\usepackage{amsmath}
\usepackage{graphicx}
\usepackage{enumerate}
\usepackage{natbib}
\usepackage{url} 
\usepackage{psfrag,epsf}
\usepackage[utf8]{inputenc} 
\usepackage[T1]{fontenc}    
\usepackage{booktabs}       
\usepackage{amsfonts}       
\usepackage{nicefrac}       
\usepackage{microtype}      
\usepackage{color}
\usepackage{amssymb}
\usepackage[ruled, vlined, linesnumbered]{algorithm2e}
\usepackage{multirow}
\usepackage{booktabs}
\usepackage{geometry}
\usepackage{float} 
\usepackage{subfigure}
\usepackage{overpic}
\usepackage[flushleft]{threeparttable}
\usepackage[english]{babel}
\usepackage{amsthm}
\usepackage{bm}
\usepackage{ulem}
\usepackage{indentfirst}
\usepackage{hyperref}
\usepackage{xr}
\hypersetup{
colorlinks=true,
linkcolor=blue,
filecolor=black,
urlcolor=red,
citecolor=blue,
}

\newtheorem{theorem}{Theorem}[section]
\newtheorem{corollary}{Corollary}[section]

\newtheorem{definition}{Definition}[section]
\newtheorem{assumption}{Assumption}[section]
\newtheorem{proposition}{Proposition}[section]

\newcommand{\blind}{1}

\newcommand{\mcH}{\mathcal{H}}
\newcommand{\mcG}{\mathcal{G}}
\newcommand{\mcL}{\mathcal{L}}
\newcommand{\mcF}{\mathcal{F}}

\newcommand{\bX}{\bm{X}}
\newcommand{\bY}{\bm{Y}}
\newcommand{\bZ}{\bm{Z}}
\newcommand{\bB}{\mathbf{B}}

\newcommand{\mbE}{\mathbb{E}}

\begin{document}

\date{}

\def\spacingset#1{\renewcommand{\baselinestretch}%
{#1}\small\normalsize} \spacingset{1}


\if1\blind
{
  \title{\bf Generative adversarial learning  with optimal  input dimension and its adaptive generator architecture}
  \author{Zhiyao Tan*, Ling Zhou\thanks{Co-first authors.} \   and  Huazhen Lin\thanks{Corresponding author. Email address: \emph{linhz@swufe.edu.cn}. } \thanks{The research was supported by National Key R\&D Program of China (No.2022YFA1003702), National Natural Science Foundation of China (Nos. 11931014 and 12271441), and  New Cornerstone Science Foundation.}\vspace{.5cm}\\
   New Cornerstone Science Laboratory, \\
    Center of Statistical Research and School of Statistics,\\
     Southwestern University of Finance and Economics, Chengdu, China}
  \maketitle
} \fi

\if0\blind
{
  \bigskip
  \bigskip
  \bigskip
  \begin{center}
    {\LARGE\bf Title}
\end{center}
  \medskip
} \fi

\bigskip
\begin{abstract}
In this paper, we investigate the impact of the input dimension on the generalization error in generative adversarial networks (GANs). In particular, we first provide both theoretical and practical evidence to validate the existence of an optimal input dimension (OID) that minimizes the generalization error. Then, to identify the OID, we introduce a novel framework called generalized GANs (G-GANs), which includes existing GANs as a special case. By incorporating the group penalty and the architecture penalty developed in the paper, the proposed G-GANs have several intriguing features.
First, our framework offers adaptive dimensionality reduction from the initial dimension to a dimension necessary for generating the target distribution.
Second, this reduction in dimensionality also shrinks the required size
of the generator network architecture, which is automatically identified by the proposed architecture penalty.
Both reductions in dimensionality and the generator network significantly improve the stability and the accuracy of the estimation and prediction.
Theoretical support for the consistent selection of the input dimension and the generator network is provided.
Third, the proposed algorithm involves an end-to-end training process, and the algorithm allows for dynamic adjustments between the input dimension and the generator network during training, further enhancing the overall performance of the G-GANs.
Extensive experiments conducted with simulated and benchmark data demonstrate the superior performance of the proposed G-GANs. In particular, compared to that of off-the-shelf methods, the proposed method achieves an average improvement of $45.68\%$ in the CT slice dataset, $43.22\%$ in the MNIST dataset and $46.94\%$ in the FashionMNIST dataset in terms of the maximum mean discrepancy or Fréchet inception distance.
Moreover, the features generated based on the input dimensions identified by G-GANs align with visually significant features: thickness and angle of inclination of the digits in MNIST and fabric quantity and clothing style in FashionMNIST.
\end{abstract}

\noindent%
\vfill

\newpage
\spacingset{1.9} 
\section{Introduction}
\label{sec:intro}

Deep generative models, such as generative adversarial networks (GANs),
have gained significant attention in recent years due to their ability to learn complex data distributions and to generate high-quality samples \citep{goodfellow2014generative,radford2015unsupervised,arjovsky2017wasserstein,karras2019style}.
GANs
are performed by training a discriminator $f$ and a generator $g$ in a confrontational fashion until the samples are generated by an easy-to-sample source distribution $\nu$
are indistinguishable from the sampling of the target distribution $\mu$. Particularly,
GANs are formulated as a minimax optimization problem at the population level:
\begin{eqnarray} \label{f1}
    g^* &=& \underset{g \in \mathcal{G}}{\arg\min} \ \underset{f \in \mathcal{F}}{\max} \ \mathcal{L}(f,g), \\
    \mathcal{L}(f,g) &=& \mathbb{E}_{x \sim \mu} [f(x)] - \mathbb{E}_{z \sim \nu} [f(g(z))],
    \label{oGAN}
\end{eqnarray}
where both the generator and the discriminator classes, denoted as $\mathcal{G}$ and $\mathcal{F}$, respectively, are parameterized by neural networks.

The underlying principle of GANs is built upon the assumption that real-world datasets possess low-dimensional intrinsic structures
\citep{arjovsky2017towards, dahal2023deep}.
Leveraging this assumption, GANs create high-dimensional data of dimension $D$ from low-dimensional input variables $z$ with a significantly smaller dimension of $d$, i.e., $d \ll D$.
However, both the theoretical and practical understanding of $d$ remain unclear. In particular, given observed data and an easy-to-sample distribution, such as a uniform or Gaussian distribution,
how the generalization error of a GAN depends on the input dimension $d$.

\begin{figure}
\centering
    \subfigure[M1]{
        \begin{minipage}[t]{0.23\linewidth}
        \centering
        \includegraphics[width=1.7in]{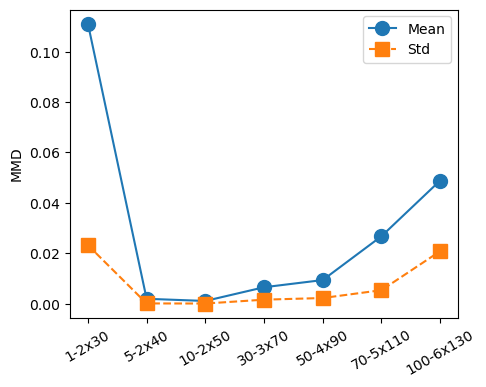}
        \end{minipage}%
    }
    \subfigure[M2]{
        \begin{minipage}[t]{0.23\linewidth}
        \centering
        \includegraphics[width=1.7in]{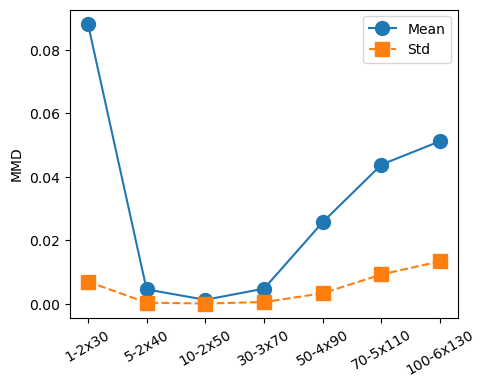}
        \end{minipage}%
    }
    \subfigure[M3]{
        \begin{minipage}[t]{0.23\linewidth}
        \centering
        \includegraphics[width=1.7in]{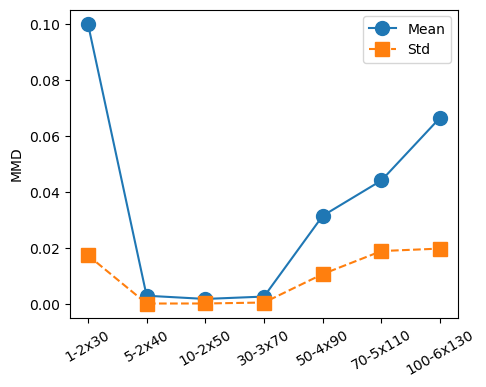}
        \end{minipage}%
    }
    \subfigure[M4]{
        \begin{minipage}[t]{0.23\linewidth}
        \centering
        \includegraphics[width=1.7in]{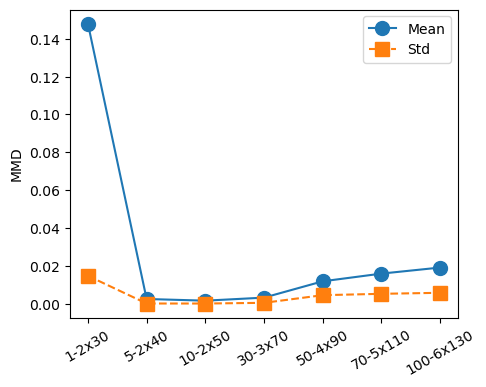}
        \end{minipage}%
    }
    \spacingset{0.95}

\caption{The mean (blue solid line) and standard deviation (Std, orange dotted line) of Maximum Mean Discrepancy (MMD)  of SNGANs \citep{miyato2018spectral} based on 10 replications   with varying input dimensions and  corresponding generator architectures, where  $d$-$l\times w$ indicates the generator with  the input dimension $d$,  depth $l$ and width  $w$. (M1)-(M4) refer to four numerical simulations,  described in detail in Section \ref{sec:exper}.
}
\label{mmd_t1}
\end{figure}

The generalization error of GANs can be decomposed into three main components: the generator approximation error, the discriminator approximation error, and the statistical error.
The generator approximation error pertains to the capacity of neural networks to approximate the target distribution.
\cite{lee2017ability} shows that the target distributions composed of Barron functions \citep{barron1993universal} can be approximated by a neural network.
However, when the target distribution is Gaussian, \cite{bailey2018size} shows that the generator approximation error takes the form $\mathcal{O}(W)^{-\mathcal{O}(L/D)}$, where $W$ and $L$ represent the width and depth of the neural network, respectively.
To relax the restriction of a Gaussian target distribution to any continuous density,
\cite{lu2020universal} establish a generator approximation rate $\mathcal{O}(W^{-\frac{1}{D}})$ of neural networks
with the requirement that $L=\mathcal{O}( \log W )$, and the same dimension for the source and the target distributions, $d$ is taken to be $D$.
This rate is further extended to target distributions with H{\"o}lder densities by \cite{chen2022distribution}.
However, all these works suffer from the curse of dimensionality due to the approximation rates $\mathcal{O}(W)^{-\mathcal{O}(L/D)}$ or $\mathcal{O}(W^{-\frac{1}{D}})$ for medium to large $D$.
Recent studies \citep{perekrestenko2020constructive, yang2022capacity, huang2022error} demonstrate that
a high-dimensional target distribution can be generated by one-dimensional distributions, and the approximation error vanishes as $L$ becomes infinite or $W \ge 7D+1$.
These methods focus on approximating the empirical distribution, and the regularity or smoothness of the target distribution is ignored, which may lead to poor generalization errors.
This scenario is illustrated in Figure \ref{mmd_t1} and supplementary Figure E2.
In Figure \ref{mmd_t1}, the leftmost side of each graph shows the generalization errors for $d=1$, which are much larger than those for $d>1$.
In supplementary Figure E2,  on the GRID dataset, the samples generated by WGAN-GP \citep{gulrajani2017improved} with $d=2$ perform much better than  those with $d=1$ due to the former capture spatial information.

The discriminator approximation error relies on the ability of the neural network to represent functional classes of $D$-dimensional data because the discriminator works on the observed data.
Recent advances in deep learning have spurred numerous studies on
quantifying the approximation error of deep ReLU networks based on the number of parameters or neurons, and the results have already been used to bound the discriminator approximation error \citep{huang2022error, chen2022distribution, yarotsky2017error, yarotsky2018optimal,  
shen2019deep, jiao2023deep},  which has order $\mathcal{O}((WL)^{-2\beta/D})$, where $\beta$ is the smoothness index of the evaluation function class.

Regarding the statistical error,
\cite{chen2022distribution} demonstrated that it scales as $\mathcal{O}(n^{-\frac{\beta}{2\beta+D}})$
when $\mathcal{F}$ belongs to a H{\"o}lder class with smoothness index $\beta$
or has a low-dimensional linear structure.
The distance induced by the H{\"o}lder class is quite general. However, these results still suffer  the curse of dimensionality.
When $D$ is large, challenges arise in explaining the performance of GANs.
In
\cite{schreuder2021statistical} and \cite{dahal2023deep}, the authors demonstrate that the statistical error converges at a rate of $\mathcal{O}(\max(n^{-{\frac{\beta}{d^*}}}, n^{-{\frac{1}{2}}}))$
if we take $d=d^*$, where $d^*$ is the intrinsic dimension.
Since prior knowledge of the intrinsic dimension is usually unavailable in practice,
\cite{block2021intrinsic} and \cite{huang2022error}
further establish the rate  without  the restriction of $d = d^*$,
which is a significant contribution to the field.

Based on existing research on the above three components, we can see that the input dimension $d$ and the intrinsic dimension $d^*$
play crucial roles in the generator approximation error and the statistical error of GANs, respectively.
Given that $d^*$ is inherently unchangeable, several interesting questions can be raised:
Can we choose a proper input dimension to minimize generalization errors? How does the generalization error of GANs vary with the input dimension?
Additionally, what network architecture should the generator $g$ have for a given input dimension? Figure \ref{mmd_t1} illustrates how the generalization error of GANs first decreases then increases with the growth of the input dimension and the corresponding generator architecture while keeping the discriminator network constant. This elucidates the presence of the optimal input dimension for GANs and indicates the need to design and to train GANs with the optimal input dimension and the corresponding generator architecture that minimizes the generalization error.
In this paper, we provide new theoretical and methodological insights into input dimensions, and we bridge the gap
between the generalization error of GANs and the input dimension to answer  the  questions above.

Our key strategy to address these problems is to incorporate a matrix $\bB$ into GANs. In particular, we generalize GANs by considering the following framework:
\begin{equation}
    \mathcal{L}(f, g, \bB) = \mathbb{E}_{x \sim \mu} [f(x)] - \mathbb{E}_{z \sim \nu} [f(g(\bB z))],
    \label{Gan}
\end{equation}
Obviously, when $\bB$ is the identity matrix, the criterion $\mathcal{L}(f, g, \bB)$ is reduced to $\mathcal{L}(f, g)$, as defined in (\ref{oGAN}). Hence, the existing GANs can be regarded as a special case of the generalized GANs (G-GANs) in (\ref{Gan}).
The introduction of $\bB$ offers us an opportunity to identify the input dimension
through distinguishing rows $\bB=(B_1,\cdots,B_d)^\prime$ as zero or nonzero.
By excluding all the $B_j$ that are zero, we can determine the number of nonzero elements in the vector $\bB z$, which, in turn, determines the input dimension.
Hence, the proposed G-GANs offer adaptive dimensionality reduction from the initial $z$ dimension to a dimension necessary to generate the target distribution.

In addition to imposing the group sparsity penalty \citep{yuan2006model}
into $\bB$, we incorporate a designed architecture penalty into the generator $g$, as shown later in Section~\ref{sec:method}. In this way, we present a powerful and adaptive estimation for the input dimension and the corresponding generator network architecture,
leading to enhanced accuracy and stability in estimation and prediction, as illustrated in Figure \ref{mmd_t1}.
Unlike traditional two-step methods that first estimate the intrinsic dimension and then design the network accordingly \citep{schreuder2021statistical}, the proposed procedure is implemented through an end-to-end training process,
and this procedure allows for dynamic adjustments between the input dimension and the generator network architecture during training, further enhancing the overall performance of the G-GANs. Comprehensive experiments on various simulated datasets, the CT slice dataset and two benchmark image datasets, MNIST and FashionMNIST, demonstrate the superior effectiveness of our approach. In particular, in comparison to the off-the-shelf models, we achieved an average improvement of $45.68\%$ in the CT slice dataset, $43.22\%$ in the MNIST dataset and $46.94\%$ in the FashionMNIST dataset in terms of the maximum mean discrepancy (MMD) or Fréchet inception distance (FID); see Tables \ref{result_num} to \ref{result_f_mnist}.
Moreover, as illustrated in Figure \ref{figure:interpretaion} and supplementary Figure E3, the generated features based on the input dimensions identified by G-GANs align with visually significant features.
For example, they represent the thickness and the angle of inclination of digits in MNIST and the fabric quantity and clothing style in FashionMNIST, implying a degree of interpretability in these input dimensions.

In addition to the superior practical performance as mentioned above, this work is the first attempt to  explicitly show the effect of the input dimension $d$ on the generalization error of GANs, particularly regarding generator approximation error, as shown in Theorem \ref{t1}.
This approach is significantly different from existing works that assume either $d = D$ or $d = 1$. For the former case, i.e., $d = D$, the established approximation rate of $\mathcal{O}(W)^{-\mathcal{O}(L/D)}$ or $\mathcal{O}(W^{-\frac{1}{D}})$ \citep{bailey2018size, lu2020universal} produces the curse of dimensionality, especially for large $D$. For the latter case, i.e., $d = 1$, the generator approximation error is overlooked due to the vanishing approximation error for the empirical target distribution. As mentioned above, approximating the empirical distribution instead of the population distribution can lead to overfitting and can result in poor generalization error, as depicted in Figure \ref{mmd_t1} and supplementary Figure E2.
In contrast to these studies exclusively focusing on either $d = D$ or $d = 1$, our derived Theorem~\ref{t1} shows how the generator approximation error and the statistical error rely on the input dimension $d$ and the size of the generated sample $m$. Particularly, when $m>n$, the generator approximation error and the statistical error are characterized by $\mathcal{O}_p(n^{\frac{-\beta_1}{2 \beta_1 + d}} + \underset{\bar{g}^d \in \mathcal{H}^{\beta_1}([0,1]^d)}{\inf}d_{\mcF}(\bar{g}^d_{\#}\nu^d, g^{d_0}_{0\#}\nu^{d_0}))$ and $\mathcal{O}_p(n^{\frac{-\beta_1}{2 \beta_1 + d}}\log^2n+n^{\frac{-\beta_2}{2 \beta_2 + d^*}}\log^2n)$, respectively, where $d_0$ is the minimal input dimension (MID) defined in Definition \ref{def:oid} and $\beta_1$ and $\beta_2$ are the smoothness indices of the H{\"o}lder class for the generator and the discriminator functions, respectively.
$d_{\mcF}(\cdot,\cdot)$ is defined by (\ref{df}).
If $d< d_0$, the term $n^{\frac{-\beta_1}{2 \beta_1 + d}}$ is well controlled, while the term $\underset{\bar{g}^d \in \mathcal{H}^{\beta_1}([0,1]^d)}{\inf}d_{\mcF}(\bar{g}^d_{\#}\nu^d, g^{d_0}_{0\#}\nu^{d_0})$ remains significantly different from zero because of the limitation of approximating a $d_0$-dimensional H{\"o}lder function by using a smaller $d$-dimensional counterpart.
When $d \ge d_0$, the generator approximation error $\underset{\bar{g}^d \in \mathcal{H}^{\beta_1}([0,1]^d)}{\inf}d_{\mcF}(\bar{g}^d_{\#}\nu^d, g^{d_0}_{0\#}\nu^{d_0})$ vanishes.
However, as the input dimension $d$ increases, a larger generator network becomes necessary, leading to an increase in the statistical error. Consequently, there
exists a trade-off between the generator approximation error and the statistical error, suggesting the existence of an optimal input dimension $d$.
This is consistent with the practical performance shown in Figure \ref{mmd_t1}.

In addition, we note that in most existing works, it is assumed that $m$ can be arbitrarily large so that the statistical error induced by $m$ can be ignored \citep{arora2017generalization,zhang2017discrimination,schreuder2021statistical, block2021intrinsic, dahal2023deep}. Only a few authors explicitly show the impact of $m$. For $d = D$, \cite{chen2022distribution} shows that when $m \geq n$, the impact of $m$ can be ignored, and the generalization error is determined by $\mathcal{O}(n^{-\frac{2\beta}{2\beta + D}})$. \cite{huang2022error} shows that for $d = 1$, only when $m \geq n^{2 + 2\beta/D}\log^6 n$ can the impact of $m$ be ignored. However, in practice, given that $n$ is already large, such an extraordinarily large requirement on $m$ is practically infeasible due to limited storage and time costs. Based on our Theorem \ref{t1}, a well-selected $d$, compact generator architecture, and $m \geq n$ are sufficient to yield superior performance in approximating the population target distribution. Thus, we further develop a novel adaptive method
to automatically determine the optimal $d$ and its corresponding generator architecture.
Theoretical support for the consistency of the input dimension and the generator architecture is provided in Theorem \ref{t2}.

The remainder of this paper is organized as follows. We begin by introducing the preliminaries and notation.
In Section \ref{sec:theorem}, we establish a theory to uncover a trade-off between the approximation error of the generator and the statistical error.
We introduce the proposed G-GANs method, and we establish the theoretical properties of the estimators, including the consistency of the input dimension and the generator architecture, in Section~\ref{sec:method}. The implementation procedure is presented in Section~\ref{sec:imple}.
In Section \ref{sec:exper},
we conduct extensive simulation studies
and apply our method to the CT slice dataset and two benchmark image datasets, MNIST and FashionMNIST.
Finally, we conclude our findings in Section \ref{sec:conc}.
We provide additional numerical experiment results and proofs
in the supplementary material.

\section{Theoretical results}
\label{sec:theorem}

In this paper, we consider a feedforward neural network (FNN) with a ReLU activation function to approximate the generator and the discriminator functions.
In particular, a ReLU neural network with $L$ hidden layers is a collection of mappings $\phi : \mathbb{R}^{N_0} \to \mathbb{R}^{N_{L+1}}$ of the form $\phi(x) = T_L \circ \sigma \circ T_{L-1} \circ \cdots \circ \sigma \circ T_0(x)$,
where $\phi_1 \circ \phi_2(x) := \phi_1(\phi_2(x))$ represents the composition of two functions $\phi_1$ and $\phi_2$. $\sigma(x):= \max(x, 0)$ is the ReLU function, which is applied elementwise; $T_l(x) := A_lx+c_l$ is an affine transformation with $A_l \in \mathbb{R}^{N_{l+1} \times N_l}$ and $c_l \in \mathbb{R}^{N_{l+1}}$ for $l=0, 1, ..., L$, and $N_l$ is the number of neurons in layer $l$.
We term $W=\max\{N_1, N_2, ..., N_L\}$ and $L$ the \textit{width} and the \textit{depth}, respectively, of the neural network. The \textit{size} $\mathcal{S}$ is defined as the total number of parameters in the network $\phi$, i.e., $\mathcal{S}=\sum_{i=0}^L N_{i+1} \times (N_i+1)$.
We denote by $\mathcal{NN}(W, L, \mathcal{S}, \mathcal{B})$ the set of ReLU neural networks $\phi$ with width $W$, depth $L$, size $\mathcal{S}$ and $\lVert \phi \rVert_{\infty} \le \mathcal{B}$ for some $0 < \mathcal{B} < \infty$, where $\lVert \phi \rVert_{\infty}$ is the supnorm of the function $\phi$.

We consider the H{\"o}lder class as the evaluation class $\mathcal{H}$ to which the true functions belong.
Let $\beta=s+r>0$, $r\in (0,1]$ and $s=\left\lfloor \beta \right\rfloor$, where $\left\lfloor \beta \right\rfloor$ denotes the largest integer strictly smaller than $\beta$. The H{\"o}lder class $\mathcal{H}^{\beta}(\mathcal{X})$ takes the form
$ \mathcal{H}^{\beta}(\mathcal{X}) := \{h : \mathcal{X} \to \mathbb{R}, \underset{\lVert \alpha \rVert_1 < s}{\max} \lVert \partial^{\alpha}h \rVert_{\infty} \le 1, \underset{\lVert \alpha \rVert_1 = s}{\max} \underset{x \neq y}{\sup} \frac{|\partial^{\alpha}h(x) - \partial^{\alpha}h(y)|}{\lVert x-y\rVert_2^r} \le 1 \},$
where $\partial^\alpha = \partial^{\alpha_1} \cdots \partial^{\alpha_d}$ with $\alpha=(\alpha_1, \cdots, \alpha_d)^\top$, $\lVert \alpha \rVert_1=\sum_{i=1}^d \alpha_i$ and $\| \alpha\|_2 = \sqrt{\sum_{i=1}^d \alpha_i^2}$. The H{\"o}lder class is broad enough to cover most applications. However,
the evaluation class $\mcH$ does not have to coincide with the generator class $\mathcal{G}$ or the discriminator class $\mathcal{F}$,
which consist of ReLU-based FNNs. For a measurable mapping $\nu$, we define the push-forward measure $g_{\#}\nu$ as $g_{\#}\nu(\mathcal{A}):=\nu(g^{-1}(\mathcal{A}))$ for a measurable set $\mathcal{A}$.

Based on a finite collection of samples $X_1, ..., X_n \overset{i.i.d}{\sim} \mu$ and $Z_1, ..., Z_m \overset{i.i.d}{\sim} \nu$, we replace $\mu$ and $\nu$ with their empirical counterparts $\mu_n=\frac{1}{n}\sum_{i=1}^n\delta_{X_i}$ and $\nu_m=\frac{1}{m}\sum_{j=1}^m\delta_{Z_j}$, respectively, and we estimate $g$ by
\begin{equation} \label{f2}
    \hat{g}
    = \underset{g \in \mathcal{G}}{\arg\min} \ \underset{f \in \mathcal{F}}{\max} \ \hat{\mathcal{L}}(f,g) : = \underset{g \in \mathcal{G}}{\arg\min} \ \underset{f \in \mathcal{F}}{\max} \left\{ \frac{1}{n}\sum_{i=1}^{n}f(X_i) - \frac{1}{m}\sum_{j=1}^m(f(g(Z_j))) \right\}.
\end{equation}

We first investigate
the generalization error of the GANs defined in (\ref{f2}).
This error is
evaluated by
the integral probability metric (IPM) \citep{muller1997integral} between the target distribution $\mu$ and the learned distribution $\hat{g}_{\#}\nu$ with respect to the evaluation class $\mathcal{H}$, which has the form of
\begin{eqnarray}
\label{df}
    d_{\mcH}(\hat{g}_{\#}\nu, \mu) = \underset{h \in \mathcal{H}}{\max} \ \mathbb{E}_{x \sim \mu}[h(x)] - \mathbb{E}_{z \sim \nu}[h(\hat{g}(z))].
\end{eqnarray}
By choosing different $\mathcal{H}$s, IPM can express many commonly used metrics \citep{arjovsky2017wasserstein,  dziugaite2015training, mroueh2017sobolev}.

Let the inequality $A \preceq B$ represent $A \leq CB$ for a positive constant $C$,
$A \wedge B=\min(A, B)$, and $\Phi_1 \circ \Phi_2:= \{
\phi_1 \circ \phi_2: \phi_1 \in \Phi_1, \phi_2 \in \Phi_2 \}$ represents the composition of two function classes $\Phi_1$ and $\Phi_2$. We denote the support set of $\mu$ as $\Omega$.
In existing works \citep{chen2022distribution, huang2022error}, an upper bound on the risk $\hat{g}$ has been obtained in terms of the evaluation class $\mcH$, which is decomposed into the following three terms (Lemma 9 in \cite{huang2022error}):
\begin{equation} \label{eq:old_decom}
    d_{\mcH}(\hat{g}_{\#}\nu, \mu) \preceq \mathcal{E}(\mathcal{H}, \mathcal{F}, \Omega) + \underset{g \in \mcG}{\inf} \ d_{\mathcal{H}}(g_{\#}\nu, \mu_n) + \{d_{\mathcal{H}}(\mu, \mu_n)\wedge d_{\mathcal{F}}(\mu, \mu_n) + d_{\mcF \circ \mcG}(\nu, \nu_m) \},
\end{equation}
where $\mathcal{E}(\mathcal{H}, \mathcal{F}, \Omega):= \underset{h \in \mathcal{H}}{\sup} \ \underset{f \in \mathcal{F}}{\inf} \ \lVert h-f \rVert_\infty$ is the discriminator approximation error from $\mathcal{F}$ to $\mathcal{H}$ on $\Omega$, which quantifies the discrepancy between the function classes $\mathcal{F}$ and $\mathcal{H}$.
The second term $\underset{g \in \mcG}{\inf} \ d_{\mathcal{H}}(g_{\#}\nu, \mu_n)$ represents the generator approximation error, which measures how well the empirical distribution $\mu_n$ can be approximated by $\nu$ through the transformation $g \in \mcG$ with respect to the metric $d_{\mathcal{H}}$.
The third term represents the statistical error, which accounts for the differences between the empirical distributions $\mu_n$ and $\nu_m$ and their population counterparts.
Notably, the upper bound in the decomposition \eqref{eq:old_decom} is not sufficiently tight to allow us to discern the impact of $\hat{g}$.
This occurs because, in the derivation, the use of the inequality $d_{\mathcal{H}}(\hat{g}_{\#}\nu_m, \mu_n) \le d_{\mathcal{H}}(g^*_{\#}\nu_m, \mu_n)$ leads to the omission of the term associated with $\hat{g}$.
To accurately calibrate the impact of learning $g$,
we derive a new decomposition via an explicit expression of $\hat{g}$ in the third term. That is,

\begin{proposition} \label{errordecomp}
Let us assume $\mathcal{F}$ is symmetric, i.e., $\phi \in \mcF$ holds if and only if $-\phi \in \mcF$.
Let $g^*$ and $\hat{g}$ be the population and the empirical GAN estimators \eqref{f1} and \eqref{f2}, respectively. Then, for any evaluation class $\mathcal{H}$, it holds that
\begin{equation} \label{bv}
        d_{\mathcal{H}}(\hat{g}_{\#}\nu, \mu) \preceq \mathcal{E}(\mathcal{H}, \mathcal{F}, \Omega) + d_{\mathcal{H}}(g^*_{\#}\nu, \mu) + \left\{d_{\mathcal{F}}(\hat{g}_{\#}\nu, \mu) - d_{\mathcal{F}}(g^*_{\#}\nu, \mu)\right\}.
    \end{equation}
\end{proposition}
The proof of Proposition~\ref{errordecomp} is straightforward by simple decomposition, and detailed proofs can be found in the Supplementary Materials B. A similar decomposition to that of \eqref{bv} can be found in \cite{zhou2022deep} for the conditional sampling problem.
The decompositions in equations~\eqref{eq:old_decom} and~\eqref{bv} differ in two ways. First, in terms of the generator approximation error in the second terms of both \eqref{eq:old_decom} and \eqref{bv}, we shift our attention from approximating the empirical distribution $\mu_n$ to approximating the population distribution $\mu$, ensuring sufficient generative capability within $\mcG$. Second, the statistical error in the third term of \eqref{bv} accounts for the discrepancy between the performance of the estimator $\hat{g}$ and the optimal possible estimator in $\mathcal{G}$, while
the statistic error in the third term of \eqref{eq:old_decom} accounts for only the differences between the empirical distributions and their population versions.

In addition to Proposition \ref{errordecomp}, we need two definitions to express the generalization error of GANs. Following \cite{schreuder2021statistical}, we assume that the true source distribution $\nu$ has been accessed and that there is a real generation function behind the target distribution.
Let us denote the true generator function $g_0$, which is supported on a bounded set; for simplicity, we assume this bounded set to be $[0, 1]^d$.
Without loss of generality, we assume $\Omega \subseteq [0,1]^D$. Since the true function may not be unique, we define:

\begin{definition} \label{a1}
A mapping class $\mathcal{G}_0:=\underset{d \le D}{\cup}\{g_0:[0,1]^{d} \to [0,1]^D, g_0 \in \mathcal{H}^{\beta_1}([0,1]^d)\}$ such that $\forall g_0 \in \mathcal{G}_0$, $\mu = g_{0\#}(\nu)$.
\end{definition}

Definition~\ref{a1} implies that
$\min_{g \in \mcG_0}\max_{h \in \mcH}\mcL(h, g) \equiv \min_{g \in \mcG_0}\max_{f \in \mcF}\mcL(f, g)$ since $\mu \equiv g_{0\#}(\nu)$ leads to $f\circ \mu \equiv f \circ g_{0\#}(\nu)$ regardless of whether $f$ belongs to $\mathcal{F}$ or $\mathcal{H}$.
Because there may be multiple different $d$s that can generate the target distribution $\mu$, we give the following definition for the minimal
input
dimension.

\begin{definition}[Minimal input dimension] \label{def:oid}
We write $g^d$ as the $d$-dimensional function and $\nu^d$ as the source distribution of the $d$-dimension. The minimal input dimension (MID) is defined as
$$d_0 := \underset{d}{\min}\{d \ | \ \mu = g^d_{0\#}(\nu^d), g_0 \in \mathcal{G}_0 \},$$
where the minimum is taken among all $d$s such that $(g^d,\nu^d)$ can exactly generate the target distribution.
\end{definition}

We now use Theorem \ref{t1} to explicitly show the upper bound of the generalization
error in terms of the sample size and the neural network architectures. Without confusing the context, we abbreviate $g^d$ and $\nu^d$ as $g$ and $\nu$, respectively.
Based on Proposition~\ref{errordecomp}, the results of approximation errors \citep{huang2022error, jiao2023deep}, the covering number on the neural network \citep{bartlett2019nearly} and by using empirical process techniques, we obtain the following theorem.

\begin{theorem} \label{t1}
    Suppose the target distribution $\mu= g_{0\#}(\nu)$,
    the true generators belong to $\mathcal{H}^{\beta_1}([0,1]^d)$ and the evaluation
    class is $\mathcal{H} = \mathcal{H}^{\beta_2}([0,1]^D) (\beta_2 \ge 1)$. Then there exists a generator class $\mathcal{G}=\{g:\mathbb{R}^d \to \mathbb{R}^D \ | \ g \in \mathcal{NN}(W_{\mathcal{G}},L_{\mathcal{G}},\mathcal{S}_{\mathcal{G}}, \mathcal{B}_{\mathcal{G}}) \}$ with
    \begin{equation} \label{eq:net-gene}
        W_{\mathcal{G}}L_{\mathcal{G}} \preceq n^{\frac{d}{2(2 \beta_1 + d)}},
    \end{equation}
    and a discriminator class $\mathcal{F}=\{f:\mathbb{R}^D \to \mathbb{R} \ | \ f \in \mathcal{NN}(W_{\mathcal{F}},L_{\mathcal{F}},\mathcal{S}_{\mathcal{F}}, \mathcal{B}_{\mathcal{F}}) \}$ with
    \begin{equation}
        W_{\mathcal{F}}L_{\mathcal{F}} \preceq n^{\frac{D}{2(2 \beta_2 + D)}},
    \end{equation}
    so that GAN estimator \eqref{f2} satisfies
    \begin{equation} \label{eq:t1_bound}
        \begin{aligned}
        d_{\mathcal{H}^{\beta_2}}(\hat{g}_{\#}\nu, \mu) & =
        \mathcal{O}_p\left(
        n^{\frac{-\beta_2}{2 \beta_2 + D}}
        + \{ n^{\frac{-\beta_1}{2 \beta_1 + d}} + \underset{\bar{g}^d \in \mathcal{H}^{\beta_1}([0,1]^d)}{\inf}d_{\mcF}(\bar{g}^d_{\#}\nu^d, g^{d_0}_{0\#}\nu^{d_0}) \} \right.\\
        &\left.\hspace{-1.5cm} + \{ n^{\frac{-\beta_1}{2 \beta_1 + d}}\log^2n
        + n^{\frac{-\beta_2}{2 \beta_2 + D}}\log^2n
        + m^{\frac{-1}{2}}n^{\frac{d}{2(2\beta_1+d)}}\log^{\frac{3}{2}}n \log^{\frac{1}{2}}m
        + m^{\frac{-1}{2}}n^{\frac{D}{2(2\beta_2+D)}}\log^{\frac{3}{2}}n \log^{\frac{1}{2}}m \}
        \right).
        \end{aligned}
    \end{equation}
\end{theorem}

Based on Theorem~\ref{t1}, we can see that the discriminator approximation error is bounded by $\mathcal{O}_p( n^{\frac{-\beta_2}{2 \beta_2 + D}})$, the generator approximation error is $\mathcal{O}_p(n^{\frac{-\beta_1}{2 \beta_1 + d}} + \underset{\bar{g}^d \in \mathcal{H}^{\beta_1}([0,1]^d)}{\inf}d_{\mcF}(\bar{g}^d_{\#}\nu^d, g^{d_0}_{0\#}\nu^{d_0}))$, and the statistical error is $\mathcal{O}_p(n^{\frac{-\beta_1}{2 \beta_1 + d}}\log^2n + n^{\frac{-\beta_2}{2 \beta_2 + D}}\log^2n + m^{\frac{-1}{2}}n^{\frac{d}{2(2\beta_1+d)}}\log^{\frac{3}{2}}n \log^{\frac{1}{2}}m + m^{\frac{-1}{2}}n^{\frac{D}{2(2\beta_2+D)}}\log^{\frac{3}{2}}n \log^{\frac{1}{2}}m)$.
The approximation error $\mathcal{O}_p(n^{\frac{-\beta_2}{2 \beta_2 + D}})$ and the statistical error $\mathcal{O}_p(n^{\frac{-\beta_2}{2 \beta_2 + D}}\log^2n)$ from the discriminator imply
that when the dimension of the observed data $D$ is large, the process suffers from the curse of dimensionality.
In recent studies \citep{nakada2020adaptive,huang2022error,block2021intrinsic}, the authors have highlighted an important insight: The intrinsic low dimensionality of observed data is the main factor that determines the performance of deep neural networks. This finding suggests that if there exists a constant $C$ and an intrinsic dimension $d^*$ ($d^* \le D$) such that $N(\epsilon, \Omega, \lVert \cdot \rVert_\infty) \le C \epsilon^{-d^*}$,
the two terms  $\mathcal{O}_p(n^{\frac{-\beta_2}{2 \beta_2 + D}})$ and $\mathcal{O}_p(n^{\frac{-\beta_2}{2 \beta_2 + D}}\log^2n)$ can be improved to $\mathcal{O}_p(n^{\frac{-\beta_2}{2 \beta_2 + d^*}})$ and $\mathcal{O}_p( n^{\frac{-\beta_2}{2 \beta_2 + d^*}}\log^2n)$, respectively, by choosing $W_{\mathcal{F}}L_{\mathcal{F}} \preceq n^{\frac{d^*}{2(2 \beta_2 + d^*)}}$, where
$N(\epsilon, \Omega, \rho)$ is the covering number of $\Omega$ under the metric $\rho$ with radius $\epsilon$.

It is interesting to compare the error bound in expression~\eqref{eq:t1_bound} with the existing results, which
either assume $d = D$ or $d = 1$. When $d=D$, our results with $\beta_1 = \beta_2 = \beta$ and $m > n$ simplify to $\mathcal{O}_p(n^{\frac{-\beta}{2 \beta + D}}\log^2n)$, which is consistent with the results in \cite{chen2022distribution}.
When $d=1$, \cite{huang2022error}
uses a specific architecture that handles the generator approximation error
attributed to the following three conditions. First, the generator size, $W^2_{\mathcal{G}}L_{\mathcal{G}} \preceq n$, is much larger than that in our proposal, i.e., $W_{\mcG}L_{\mcG} \preceq n^{\frac{d}{2(2\beta_1 + d)}}$. Second, their approach, based on memorizing empirical data, overlooks the smoothness of the target distribution, leading to poor generalizability.
Finally, their theoretical requirement $m > n^{2+2\beta/d}\log^6n$ (Theorem 5 in \cite{huang2022error}) is impractical. For instance, with $n = 10$ thousand, $m$ must reach approximately $100$ million, incurring intensive economic and time costs. More aggressively, in most existing studies the authors assume that $m$ can be arbitrarily large, which indicates that the generator is actually obtained via
\[
\hat{g}_{n} = \arg\min_{g \in \mcG}\max_{f \in \mcF} \left( \frac{1}{n}\sum_{i=1}^n f(\bX_i) - \mbE(f(g(\bZ_j))) \right) :=
\arg\min_{g \in \mcG}\max_{f \in \mcF} \tilde{\hat{\mcL}}(f, g).
\]
In practice, for any realization $m$, we have two additional terms: the last two terms in equation~\eqref{eq:t1_bound} for
the statistical errors of the generator and discriminator based on $m$ generated samples. Clearly, as long as $m > n$, the last two terms are dominated by the former two terms
$n^{\frac{-\beta_1}{2 \beta_1 + d}}\log^2n + n^{\frac{-\beta_2}{2 \beta_2 + D}}\log^2n$ in \eqref{eq:t1_bound}, indicating that the term induced by $m$ samples can be ignored if $m>n$.

Theorem \ref{t1} implies that a well-selected $d$, compact architecture, and reasonable $m$ can yield superior performance in approximating the population target distribution.
Specifically, Theorem~\ref{t1} suggests that the choice of the input dimension $d$ can affect the convergence rate by influencing the generator approximation error and the statistical error.
The generator approximation error first decreases then increases as the input dimension $d$ increases. Particularly, when $d < d_0$, the first term $n^{\frac{-\beta_1}{2 \beta_1 + d}}$ is well controlled, while the second term $\underset{\bar{g}^d \in \mathcal{H}^{\beta_1}([0,1]^d)}{\inf}d_{\mcF}(\bar{g}^d_{\#}\nu^d, g^{d_0}_{0\#}\nu^{d_0})$ remains significantly distant from zero because of the
limitation of approximating a $d_0$-dimensional H{\"o}lder function by using a smaller $d$-dimensional counterpart.
When $d \ge d_0$, the term $\underset{\bar{g}^d \in \mathcal{H}^{\beta_1}([0,1]^d)}{\inf}d_{\mcF}(\bar{g}^d_{\#}\nu^d, g^{d_0}_{0\#}\nu^{d_0})$ or the generator approximation error vanishes.
However, as the input dimension $d$ increases, according to the inequality~\eqref{eq:net-gene}, the width and depth of the required generator architecture increase:
A larger generator network becomes necessary, leading to an increase in the statistical error. Consequently, there
exists a trade-off between the generator approximation error and the statistical error, suggesting the existence of an optimal input dimension $d$.
By choosing the optimal input dimension for $d$ and its corresponding generator architecture, a faster convergence rate can be achieved for the GAN estimator.
Based on Theorem~\ref{t1}, we can directly obtain the following corollary to characterize the optimal input dimension:
\begin{corollary}
Under the conditions of Theorem~\ref{t1}, if $m > n$, the GAN estimator \eqref{f2} achieves the optimal error rate with $d = d_0$ on the order of
\[
 d_{\mathcal{H}^{\beta_2}}(\hat{g}_{\#}\nu, \mu)  =
        \mathcal{O}_p\left(
        n^{\frac{-\beta_2}{2 \beta_2 + D}} \log^2 n
        +  n^{\frac{-\beta_1}{2 \beta_1 + d_0}}\log^2n \right).
\]
\end{corollary}
Hence, $d_0$ is the optimal input dimension (OID) that minimizes the generalization errors.
In the next section, we develop a novel adaptive method for choosing $d_0$ and the generator architecture simultaneously.

\section{Generalized GANs}
\label{sec:method}

As mentioned in Section \ref{sec:intro}, a key strategy for identifying the OID is to introduce an index matrix $\mathbf{B}$, leading to the generalized GANs (G-GANs) framework (\ref{Gan}). Since the row of indices $\mathbf{B}$ determines the input dimension,
determining the OID is equivalent to identifying nonzero rows of $\mathbf{B}$, which can be automatically implemented by
incorporating a group sparsity penalty on $\mathbf{B}$.
Moreover, according to Theorem~\ref{t1},
the size and the architecture of the generator should be adaptively adjusted by the OID, which is implemented by
removing redundant hidden layers, by reducing the width  and by forcing the redundant parameters to 0.
Here, we note that, as long as the product of the width and depth of the network architecture satisfies the inequality~\eqref{eq:net-gene} in Theorem~\ref{t1},
the optimal rate is achievable. That is, we can either select a proper depth with a fixed width or change the width with a fixed depth.
Since a shallow network with a large width often demands less optimization effort than does
a deep network with a small width
\citep{DBLP:journals/jmlr/GlorotB10, srivastava2015training, he2016deep}, we hence prespecify the width with a large constant. Then, we determine the generator architecture by appropriately selecting its depth.
In practical deep neural network (DNN) implementations, the depth is usually given in advance, and changing the depth requires retraining a new network.
This limitation hinders
the dynamic adjustment between $d$ and the corresponding depth of the generator architecture.
To address this,
we propose a new strategy based on the key observation that if layer $l$ is redundant,
the affine transformation $T_l(x)= A_lx+c_l$ is not necessary; that is, the information from $T_l(x)$ equals that from $x$. As a result, $A_l$ can be replaced by an identity matrix and $c_l=0$,
which is realized by
imposing a depth penalty $\|A_l -  \mathbf{I} \|_1+\|c_l\|_1$, where
$\mathbf{I}$ is an identity matrix with the same dimensions as $A_l$.

Let us assume that the generative class $\mathcal{G}$ is parameterized by a ReLU neural network $\mathcal{NN}(W_{\mathcal{G}},L_{\mathcal{G}},\mathcal{S}_{\mathcal{G}},\mathcal{B}_{\mathcal{G}})$ with parameters $\theta:=\{A_l, c_l\}_{l=0}^{L_\mcG}$ and that the discriminator class $\mathcal{F}$ is parameterized by $\mathcal{NN}(W_{\mathcal{F}},L_{\mathcal{F}},\mathcal{S}_{\mathcal{F}},\mathcal{
B}_{\mathcal{F}})$ with  parameters  $w$.
Denote  $\mathcal{W}_\mathbf{B}=\{\mathbf{B}:\mathbf{B}\in \mathbb{R}^{d \times d}, \lVert \mathbf{B} \rVert_\infty \le \kappa_{\mathcal{W}}\}$
with the $i-$th row of a matrix $\mathbf{B}$ denoted by $\mathbf{B}^{[i,:]}=(\mathbf{B}^{[i,j]}, j=1,\cdots,  d)^T$.
Then, we consider the following objective for G-GANs:
\begin{equation}\label{f8}
    \begin{aligned}
        \left( \hat{\mathbf{B}}, \hat{\theta} \right) =
       \underset{\mathbf{B} \in \mathcal{W}_{\mathbf{B}}, \theta: g_{\theta} \in \mathcal{G}}{\arg\min} \
       \underset{w:f_w \in \mathcal{F}}{\max}
       & \left\{ \frac{1}{n}\sum_{i=1}^n f_w (\bX_i) -
       \frac{1}{m}\sum_{j=1}^m f_w (g_{\theta}(\mathbf{B}\bZ_j)) + \mathcal{L}_{reg}(\mathbf{B}, \theta)
       \right\},
   \end{aligned}
\end{equation}
where $\mathcal{L}_{reg}(\mathbf{B}, \theta) = \lambda_{1} M(\mathbf{B}) + \lambda_{2}P(\theta) + \lambda_3 Q(\theta)$,
$M(\mathbf{B}) = \sum _i\lVert \mathbf{B}^{[i,:]} \rVert_2 = \sum_i \sqrt{\sum_j {\mathbf{B}^{[i,j]}}^2}$ is the group sparsity penalty on the $\bB$th row,
$P(\theta) = \sum_{l=1}^{L_\mcG-1}\|A_l -  \mathbf{I} \|_1 + \| c_l \|_1$ and $Q(\theta) = \|\theta\|_1$ are the architecture penalties, including the depth and the sparsity penalties.
$\lambda_{1}$, $\lambda_{2}$ and $\lambda_{3}$ are hyperparameters.
As $\lambda_{1}$, $\lambda_{2}$ and $\lambda_{3}$ increase, the number of nonzero rows in $\mathbf{B}$, the depth of the generator architecture and the number of nonzero components in $\theta$ decrease, resulting in a lower input dimension and a simpler neural network architecture.
Hence, the automatic selection of $d$ and the generator architecture can be accomplished by tuning the parameters $\lambda_{1}$, $\lambda_{2}$ and $\lambda_{3}$; see Section \ref{sec:imple} for details.
Let us denote the resulting estimators by $\hat{d}$ and $\hat{\gamma} = (\hat{\mathbf{B}}, \hat{\theta})$ for the number of nonzero rows of $\hat{\mathbf{B}}$ and $\gamma = (\mathbf{B}, \theta)$, respectively.

Next, we theoretically demonstrate that G-GANs with the objective function \eqref{f8} are capable of selecting the OID and the corresponding generator network in terms of selection consistency.
The following notation is needed to establish the theory. Without loss of generality, we partition $\mathbf{B}$ into two parts, $\mathbf{U} \in \mathbb{R}^{d_0 \times d}$ and $\mathbf{V} \in \mathbb{R}^{(d-d_0) \times d}$, i.e., $\mathbf{B} = [\mathbf{U}^\top, \mathbf{V}^\top]^\top$. Then, to demonstrate the consistency of the selection of the input dimension, 
show that $\lVert \hat{\mathbf{U}}^{[i,:]} \rVert_2 \neq 0$ for $i = 1, ..., d_0$, and $\lVert \hat{\mathbf{V}}^{[j,:]} \rVert_2 = 0$ for $j = 1, ..., d-d_0$ with probability $1$.
as $n,m \to \infty$.
Since the depth and sparsity depend on the width, we
denote $\Theta^*(W) = \{ \theta^*:(\mathbf{B}^*,\theta^*) \in \underset{\mathbf{B} \in \mathcal{W}_{\mathbf{B}}, \theta:g_\theta \in \mcG }{\arg \min} \ \{d_{\mcF}(g_{\theta} \circ \mathbf{B}_{\#} \nu, \mu), \mathbf{B} = \left[\mathbf{U}^\top, \bm{0}^\top\right]^{\top}, \mathbf{U} \in \mathbb{R}^{d_0 \times d}, A_l\in \mathbb{R}^{W \times W}, 1 \le l \le L_\mcG
\} \}$ for  the
 parameters set of generator with the width $W$ under the OID.

The minimal depth is then denoted by $l_\theta (W) = \underset{l^*}{\min}\{l^*: l^* = \text{dep}(\theta^*), \theta^* \in \Theta^*(W)\}$, where $\text{dep}(\theta^*)$
is the
depth of $\theta^*$.
Hereafter, we drop $W$ for notational simplicity without confusion.
Then, the estimation of the generator achieves consistency in depth selection
if $\mathbb{P}(\text{dep}(\hat{\theta})=l_\theta) \to 1$ as $n, m \to \infty$.
For the size of generator,
we denote $n_\theta = \underset{\breve{\theta}^* \in \breve{\Theta}^*}{\min} \ \lVert \breve{\theta}^* \rVert_0$ as the minimal size of the generator under the
OID and
minimal depth, where
$\breve{\Theta}^* = \{ \breve{\theta}^*: \text{dep}(\breve{\theta}^*) = l_\theta, \breve{\theta}^* \in \Theta^* \}$ is the
parameter set of the generator under the OID and
minimal depth. Notably, although $\breve{\theta}^*$ satisfying $\lVert \breve{\theta}^* \rVert_0 = n_\theta$ is not unique due to the unidentifiability of the neural network, $n_\theta$ is unique. Then, the size selection consistency for the generator network is established by showing that $\mathbb{P}(\lVert \hat{\theta} \rVert_0 = n_\theta) \to 1$,
as $n,m \to \infty$. Overall, we denote the parameter set of neural networks with OID,
minimal depth and
minimal size  by $\Gamma^* = \{ \gamma^* = (\mathbf{B}^*, \theta^*):
(\mathbf{B}^*, \theta^*) \in \underset{\mathbf{B} \in \mathcal{W}_{\mathbf{B}}, \theta: g_\theta \in \mcG}{\arg \min} \ \{d_{\mcF}(g_{\theta} \circ \mathbf{B}_{\#} \nu, \mu),  \mathbf{B}^* = \left[\mathbf{U}^\top, \bm{0}^\top\right]^{\top}, \mathbf{U} \in \mathbb{R}^{d_0 \times d}, A_l\in \mathbb{R}^{W \times W},  1 \le l \le L_\mcG , \text{dep}(\theta^*)=l_\theta, \| \theta^*\|_0 = n_\theta \} \}$. Let $d(\hat{\gamma},\Gamma^*) := \underset{\gamma^* \in \Gamma^*}{\min}\lVert \hat{\gamma} - \gamma^* \rVert_2$, and
we need the following two assumptions to establish consistency in the selection.
\begin{assumption} \label{cond:L1toL0}
For any ${\breve{\theta}}^*_1, {\breve{\theta}}^*_2 \in \breve{\Theta}^*$, $\| {\breve{\theta}}^*_1 \|_1 \le \| {\breve{\theta}}^*_2\|_1$ implies $\| {\breve{\theta}}^*_1 \|_0 \le \| {\breve{\theta}}^*_2\|_0$.
\end{assumption}

\begin{assumption} \label{cond:Bound}
For any $\tilde{\theta}^*_1, \tilde{\theta}^*_2 \in \tilde{\Theta}^*$, there exists a constant $M_b < \infty$ such that $\| \tilde{\theta}^*_1 - \tilde{\theta}^*_2\|_2 \le M_b$ where $\tilde{\Theta}^* = \{ \tilde{\theta}^*: (\tilde{\mathbf{B}}^*,\tilde{\theta}^*) \in \underset{\mathbf{B} \in \mathcal{W}_{\mathbf{B}}, \theta:g_\theta \in \mcG }{\arg \min} \ d_{\mcF}(g_{\theta} \circ \mathbf{B}_{\#} \nu, \mu) \}$.
\end{assumption}

Assumption \ref{cond:L1toL0} concerns the connection of the $L_1$ norm and the $L_0$ norm for the parameters within $\breve{\Theta}^*$.
Based on the four numerical simulation datasets outlined in Section \ref{sec:exper}, the empirical analysis presented in Supplementary Figure E1 illustrates a monotonic relationship between the $L_1$ norm and the $L_0$ norm, providing support for Assumption \ref{cond:L1toL0}.
This difference may be attributed to the relatively small magnitudes of all the neural network parameters \citep{bellido1993backpropagation, huang2021rethinking}.
Assumption \ref{cond:Bound} focuses on the bounded difference between any two parameters in $\tilde{\Theta}^*$.
This assumption is a relaxation of the assumption that
the $L_1$ norm of
any
parameter is bounded, a condition frequently required in the neural network literature \citep{ chen2022distribution}.

\begin{theorem} \label{t2}
    Considering that the generator class $\mathcal{G}=\{g:\mathbb{R}^d \to \mathbb{R}^D \ | \ g \in \mathcal{NN}(W_{\mathcal{G}},L_{\mathcal{G}},\mathcal{S}_{\mathcal{G}}, \mathcal{B}_{\mathcal{G}}) \}$ with
    \begin{equation}
        W_{\mathcal{G}}L_{\mathcal{G}} \preceq n^{\frac{d}{2(2 \beta_1 + d)}},
    \end{equation}
    and the discriminator $\mathcal{F}=\{f:\mathbb{R}^D \to \mathbb{R} \ | \ f \in \mathcal{NN}(W_{\mathcal{F}},L_{\mathcal{F}},\mathcal{S}_{\mathcal{F}}, \mathcal{B}_{\mathcal{F}}) \}$ with
    \begin{equation}
        W_{\mathcal{F}}L_{\mathcal{F}} \preceq n^{\frac{D}{2(2 \beta_2 + D)}},
    \end{equation}
    suppose that  Assumptions \ref{cond:L1toL0} and \ref{cond:Bound} hold. Let $\hat{\gamma}$ be the estimator of \eqref{f8}, when $m>n$,
    if $\lambda_1 = o(1)$, $\lambda_2 = o(\lambda_1)$, $\lambda_3 = o(\lambda_2)$,  $(n^{\frac{-\beta_1}{2 \beta_1 + d}} + n^{\frac{-\beta_2}{2 \beta_2 + D}})\log^2n=o(\lambda_3)$,
    we deduce that
    \begin{equation}
        d(\hat{\gamma}, \Gamma^*) = o_p(1).
    \end{equation}
\end{theorem}

The detailed proof of Theorem \ref{t2} can be found in Supplementary C. The entire proof includes four parts: convergence to the optimal parameter set, input dimension consistency, depth and size selection consistency for the generator, where the first is achieved by the Lojasiewicz inequality \citep{ji1992global, colding2014lojasiewicz} and Young's inequality. The second is established by
following a similar idea as those in \cite{dinh2020consistent0, dinh2020consistent},
and the last two parts can be represented by the unidentifiability of the neural network,
and the relationship between the $L_1$ penalty and $L_0$ penalty in $\breve
{\Theta}^*$ based on Assumption \ref{cond:L1toL0}.

Theorem \ref{t2} shows that G-GANs can
identify the OID and the corresponding generator architecture
with appropriate choices of $\lambda_{1}$, $\lambda_{2}$ and $\lambda_3$.
The conditions $\lambda_2=o(\lambda_1)$ and $\lambda_3 = o(\lambda_1)$ are required because the generator architecture relies on the chosen input dimension. That is, we need to determine the input dimension before identifying the generator architecture.
The condition $\lambda_3 = o(\lambda_2)$ implies that the depth should be identified before determining the sparse structure of the generator architecture.
These conditions are natural for implementing our strategy for identifying the structure of the generator.
The requirement that $\lambda_{1}$, $\lambda_{2}$ and $\lambda_3$ exceed the statistical error $(n^{\frac{-\beta_1}{2 \beta_1 + d}} + n^{\frac{-\beta_2}{2 \beta_2 + D}})\log^2n$ is to prevent cases where redundant dimensions or parameters fail to converge to 0 due to randomness.

\section{Implementation}
\label{sec:imple}

In this section, we present the detailed implementation of G-GANs.
We adopt the mini-batch stochastic gradient descent method \citep{bottou2010large, kingma2014adam} to find the solution to \eqref{f8}.
There are two technical  issues in implementing
the algorithm: the selection of appropriate penalty parameters and optimizing the neural network parameters with the $L_1$ penalty.

To mitigate bias stemming from penalties, we allow
$\lambda_1, \lambda_2, \lambda_3$  to change as iteration continues. Particularly,
we start with small penalty parameters, and we gradually increase them before subsequently decreasing them. The strategy is considered because the parameters or vectors are quite random in early iterations and then become stable as iteration continues. In addition, this approach allows us to first identify the parameters or vectors that are closer to 0 with a smaller tuning parameter, resulting in a reduction in bias from the penalty terms.
Once a significant penalty level is reached where redundant parameters are removed, we then gradually reduce the tuning parameters so that the bias for nonzero components can be well controlled. To implement these strategies, we introduce interval parameters $\Delta$, expansion factors $\delta_1$, and shrinkage factors $\delta_2$. In the first half of the training process, all penalty parameters increase by a factor of $\delta_1$ every $\Delta$ iterations. In the latter half, the parameters are decreased by a factor of $\delta_2$ every $\Delta$ iterations. For the initial penalty parameters $\lambda^{(0)}_r, r = 1, 2, 3$,
since $\lambda_1$ is used to identify the OID, while $\lambda_2$ and $\lambda_3$ are related to tuning the optimal generator architecture corresponding to the input dimension,
we consider the sequential
selection of  $\lambda^{(0)}_{1}$, $\lambda^{(0)}_2$ and $\lambda^{(0)}_3$ via grid search in some region around zero; that is,  $\lambda^{(0)}_{1}$ is first chosen such that
$\lambda_{2}$ and $\lambda_3$ are fixed at 0; then, $\lambda^{(0)}_2$ is selected with $\lambda_1$ fixed at the selected value and $\lambda_{3}$ fixed at 0. Finally, $\lambda^{(0)}_3$ is determined with $\lambda_1$ and $\lambda_2$ fixed at the selected values.

Optimizing \eqref{f8} by directly using gradient-based methods can be challenging because the penalty terms $\lVert . \rVert_{1,2}$ and $\lVert . \rVert_1$ are both nondifferentiable at $0$. To overcome this difficulty, we employ a subgradient descent algorithm \citep{boyd2004convex} with parameter truncation. The details of subgradient can be found in Supplementary D.1. To perform parameter truncation, we introduce two additional hyperparameters, $\tau_1$ and $\tau_2$, as thresholds to truncate the row $\hat{\mathbf{B}}$ with a small $L_2$ norm and the element $\hat{\theta}$ with a small absolute value, respectively. The value of $\tau_1$ can be determined
by identifying the point at which there is a significant decrease in the performance of the model after
clipping.
Following \cite{scardapane2017group},
we set $\tau_2=0.01$ in the numerical simulation and $\tau_2=0.001$ in the real image datasets. Algorithm \ref{alg_pta} presents the truncation procedure for $\mathbf{B}$ with $\tau_1$. Similarly, we can obtain
the truncation algorithm for $\theta$ with $\tau_2$.

\begin{algorithm}[H]\label{alg_pta}
\normalem
\spacingset{1}
\KwIn{$\tau_1$, $\mathbf{B}$}
\KwOut{$\mathbf{B}_{trunc}$ }
Initialization $\mathbf{B}_{trunc} \leftarrow \mathbf{B}$\;
\For{i=1:d}{
\If{$\lVert \mathbf{B}_{trunc}^{[i,:]} \rVert_2 \le \tau_1 $}{
$ \mathbf{B}_{trunc}^{[i,:]} \leftarrow 0$;
}
}
Return $\mathbf{B}_{trunc}$.
\caption{Parameter truncation algorithm}
\end{algorithm}

We iterate between updating the generator once and updating the discriminator $k$ times \citep{goodfellow2014generative, arjovsky2017wasserstein}, which are called critical steps.
The entire training process is outlined in Algorithm \ref{alg_ggan}.

\begin{algorithm}[H] \label{alg_ggan}
\normalem
\spacingset{1}
    \KwIn{$\mu_n = \{X_i\}_{i=1}^{n}$, $\nu_m =\{Z_j\}_{j=1}^{m} $}
    \KwOut{$\hat{\theta}$, $\hat{\mathbf{B}}$, $\hat{w}$}
    Initialization $\theta$, $\mathbf{B}$, $w$\;
    \For{number of training iterations}{
        \If{interval $\Delta$ step and in the first half of training iterations}{
increase the penalty parameters by a factor of $\delta_1$\;
}
\ElseIf{interval $\Delta$ step}{
decrease the penalty parameters by a factor of $\delta_2$\;
}
\For{$k$ steps}{
Sample mini-batch of $b$ samples $\{Z_1, ..., Z_b\}$ from source distribution $\nu_m$\;
Sample mini-batch of $b$ samples $\{X_1, ..., X_b\}$ from the empirical target distribution $\mu_n$\;
Update the discriminator by ascending its stochastic gradient:
$\nabla_w \ \frac{1}{b}\sum_{i=1}^{b}[f_w(X_i) - f_w(g_{\theta}(\mathbf{B}Z_i))];$

}
Sample mini-batch of $b$ samples $\{Z_1, ..., Z_b\}$ from source distribution $\nu_m$\;
Update the generator by descending its stochastic gradient: $\nabla_{\theta, \mathbf{B}} \ \frac{1}{b}\sum_{i=1}^{b} - f_w(g_{\theta}(\mathbf{B}Z_i)) + \mathcal{L}_{reg}(\mathbf{B}, \theta)$,
where
$\mathcal{L}_{reg}(\mathbf{B}, \theta) = \lambda_{1}\lVert \mathbf{B} \rVert_{1,2}  + \lambda_2 (\sum_{l=1}^{L_\mcG-1}\|A_l -  \mathbf{I} \|_1 + \| c_l \|_1)+ \lambda_{3}\| \theta\|_1$;
\If{truncating $\mathbf{B}$}{
Truncating $\mathbf{B}$ according to Algorithm \ref{alg_pta}\;
}
}
Return $\hat{\theta}$, $\hat{\mathbf{B}}$, $\hat{w}$.
\caption{Minibatch stochastic gradient descent training of G-GANs}
\end{algorithm}

\section{Experiments}
\label{sec:exper}

In this section, we investigate the performance of the proposed G-GANs
by comparing it with the existing GANs. We evaluate the ability of G-GANs in
identifying the OID and the corresponding generator
network
in various scenarios. This
includes simulated data with diverse distributions,
the CT slice dataset \citep{dua2019uci}
and two benchmark datasets:
MNIST \citep{deng2012mnist} and FashionMNIST \citep{xiao2017/online}.

The smoothing index $\beta_2$ of evaluation class $\mcH^{\beta_2}([0,1]^D)$ in Theorem \ref{t1} is usually chosen as $\beta_2 = 1$, which induces the well-known Wasserstein distance on $[0,1]^D$ \citep{villani2009optimal, huang2022error} and then Wasserstein GANs \citep{arjovsky2017wasserstein}.
Commonly used
methods
to implement Wasserstein GANs
include WGAN-GP \citep{gulrajani2017improved} and SNGAN \citep{miyato2018spectral}.
Thus, we
performed our G-GANs based on
WGAN-GP and SNGAN,
denoted as G-GAN$^W$ and G-GAN$^{SN}$, respectively.
The implementations of G-GAN$^W$ and G-GAN$^{SN}$ are similar to those of Algorithm \ref{alg_ggan}; for additional details, please refer to Algorithm D1 and Algorithm D2 in the Supplement D.2.
In addition, we also
consider several variants of G-GAN$^W$ and G-GAN$^{SN}$ to investigate the impact of the penalties in \eqref{f8}.
Specifically, we consider G-GAN$^{W\dagger}$, G-GAN$^{W\ddagger}$, G-GAN$^{SN\dagger}$ and G-GAN$^{SN\ddagger}$, where G-GAN$^{W\dagger}$ and G-GAN$^{SN\dagger}$
solely focuses on the sparsity penalty, i.e.,
$\mathcal{L}_{reg}(\mathbf{B}, \theta) = \lambda_3 \| \theta \|_1$,
and the G-GAN$^{W\ddagger}$ and G-GAN$^{SN\ddagger}$
simultaneously consider the selection of the input dimension and sparse structure,
i. e.,
$\mathcal{L}_{reg}(\mathbf{B}, \theta) = \lambda_{1}\lVert \mathbf{B} \rVert_{1,2}  + \lambda_3 \| \theta \|_1$.
For clarity and convenience, we use G-GANs$^\dagger$ to denote G-GAN$^{W\dagger}$ and G-GAN$^{SN\dagger}$ and G-GANs$^\ddagger$ for G-GAN$^{W\ddagger}$ and G-GAN$^{SN\ddagger}$.

We evaluate the performance of the methods by comparing the empirical distribution $\hat{\mu}_N = \frac{1}{N}\sum_{i=1}^N\delta_{\hat{\bX}_i}$ generated by G-GANs with the true empirical distribution $\mu_M = \frac{1}{M}\sum_{j=1}^M\delta_{{\bX}_j}$ in terms of
maximum mean discrepancy (MMD) \citep{li2015generative, dziugaite2015training}
and the Fréchet inception distance (FID) \citep{heusel2017gans}, defined as
\begin{eqnarray*} \label{def:mmd}
            \text{MMD}^2(\hat{\mu}_N, \mu_M) &=& \| \mathbb{E}_{\hat{\bX} \sim \hat{\mu}_N} \varphi(\hat{\bX}) - \mathbb{E}_{\bX \sim \mu_M} \varphi(\bX) \|_2^2 \nonumber\\
        & =& \frac{1}{N^2} \sum_{i=1}^N \sum_{i^\prime =1}^N k(\hat{\bX}_i, \hat{\bX}_{i^\prime}) + \frac{1}{NM} \sum_{i=1}^N \sum_{j =1}^M k(\hat{\bX}_i, \bX_j) + \frac{1}{M^2} \sum_{j=1}^M \sum_{j^\prime =1}^M k(\bX_j, \bX_{j^\prime}),\\
        \text{FID}(\hat{\mu}_N, \mu_M)& =& \| m_{\hat{\mu}_N} - m_{\mu_M} \|_2^2 + Tr(\Sigma_{\hat{\mu}_N} + \Sigma_{\mu_M} - 2(\Sigma_{\hat{\mu}_N} \Sigma_{\mu_M} )^{1/2}),
    \end{eqnarray*}
where $\varphi(\cdot)$ is the feature function and $k(\cdot,\cdot)$ is the kernel function for calculating the inner products.
of $\varphi(\bX)$; $\{m_{\hat{\mu}_N}, \Sigma_{\hat{\mu}_N} \}$ and $\{m_{\mu_M},  \Sigma_{\mu_M}\}$ are the mean and covariance matrix of features extracted by the inception model \citep{szegedy2015going} from $\{\hat{\bX}_i\}_{i=1}^N$ and $\{ \bX_j \}_{j=1}^M$, respectively; and $Tr(\cdot)$ is the trace operator.
To overcome tedious parameter adjustment, following
\cite{li2015generative}, we use a mixture of $K$ kernels spanning multiple ranges, i.e., $k(x,y)=\sum_{j=1}^K k_{\sigma_j}(x,y)$, and choose $\{\sigma_j\}_{j=1}^{K=3} = \{1,5,10\}$, where $k_\sigma(x,y) =\exp(-\frac{1}{2\sigma} \| x-y\|^2)$.
Typically, the FID measures the dissimilarity between two images, whereas the MMD assesses the distinction between two vectors.

The optimization algorithm, initial values for networks and source distribution are the same for all methods involved in the numerical simulations and real data. In particular,
we use the stochastic gradient descent algorithm Adam \citep{kingma2014adam} to train both the generator and the discriminator with first- and second-order momentum parameters of (0, 0.9). The generator parameters $\{A_l\}_{l=0}^{L_\mcG}$ and $\mathbf{B}$ are initialized by the normal distribution $\mathcal{N}(0, 0.004)$, and $\{c_l\}_{l=0}^{L_\mcG}$ is initialized to 0, which is the same as the discriminator parameters. The source distribution is set as Gaussian.
Parameters settings in the different datasets are summarized in Table \ref{tabel:parameter}.
All computations are implemented via PyTorch \citep{NEURIPS2019_9015} and Numpy \citep{harris2020array} in Python.

\begin{table}[]
\centering
\spacingset{0.95}
\caption{Some common parameters settings for simulations and real data in implementation. (M1)-(M4) refer to four numerical simulations which are described in detail in Section \ref{sec:numerical_simulation}. The architectures of generator and discriminator are denoted as $l \times w$, where $l$ and $w$ are the depth and width of network. The weight of gradient penalty applied only to the methods based on WGAN-GP.}
\label{tabel:parameter}
\begin{tabular}{@{}c|c|c|c@{}}
\toprule
Parameter name              & (M1)-(M4)          & CT Slices          & MNIST \& FashionMNIST \\ \midrule
Generator architecture      & $4 \times 90$      &                    & $3 \times 256$        \\
Discriminator architecture  & $4 \times 64$      &                    & $3 \times 256$        \\
Initial input dimension     & 50                 &                    & 64                    \\
Learning rate               & $2 \times 10^{-4}$ & $2 \times 10^{-4}$ & $2 \times 10^{-4}$    \\
Critical step               & 5                  & 5                  & 1                     \\
Training batch size         & 512                & 512                & 256                   \\
The weight of gradient penalty         & 10                & 5                & 10                   \\
The number of updates       & 20,000             & 20,000             & 50,000                \\
Expansion factor $\delta_1$ & 1.1                & 1.1                & 1.1                   \\
Shrinkage factor $\delta_2$ & 0.9                & 0.9                & 0.9                   \\
Interval step $\Delta$      & 100                & 100                & 250                   \\ \bottomrule
\end{tabular}
\end{table}

\subsection{Numerical simulation}
\label{sec:numerical_simulation}

We start by generating the vector $\bZ$ from a 10-dimensional standard normal distribution. We obtain 10,000 training samples and 2,000 testing samples from the following four models.

\begin{enumerate}
\item[(M1)] Linear model. Let $W$ be an $100 \times 10$ matrix with the elements $10(j-1)$ to $10j$ in the $j$th column being $[-1, -0.78, -0.56, -0.33, -0.11, 0.11, 0.33, 0.56, 0.78, 1]$,
and zero for the remaining components in the $j$th column.
We generate $\bX$ by $\bX = W\bZ$.

\item[(M2)] A two-layer rectified linear unit (ReLU) neural network model with sparse connections.
Let $W_1$ be an $50 \times 10$ matrix where the elements $(i, j)$ are $[-1, -0.5, 0, 0.5, 1]$ as $i$ varies from $5(j-1)$ to $5j$, and the remaining elements in the $j$-th column of $W_1$ are all 0.
Let us denote $W_2$ as an $100 \times 50$ matrix where nonzero values appear in rows $2j-1$ to $2j$ of
column $j$, arranged cyclically by using
the sequence $[-1, -0.78, -0.56, -0.33, -0.11, 0.11, 0.33, 0.56, 0.78, 1]$.
Then, we generate $\bX$ by $\bX = W_2\sigma(W_1\bZ)$, where $\sigma(\cdot)$ is the ReLU function.

\item[(M3)] Nonlinear model I.
The data are generated by $\bX= [(\bY[1:20]^{T})^2/4, \bY[21:50]^{T}, exp(\bY[51:70]^{T}), sin(\bY[71:100]^{T} \times 20)]^T$, where $\bY = W\bZ$ and $W$ are defined in (M1).

\item[(M4)] nonlinear model II.
The setting is similar to that of M3 except that we generated the data by $\bX = [ \sqrt{|\bY[1:20]^{T}|} - 0.1, \bY[21:50]^{T}, \log(\bY[51:70]^{T})+0.5, \cos(\bY[71:100]^{T} \times 20)]^T.$
\end{enumerate}

For models (M1)-(M4), all the network architectures, initial values, learning rates, critical steps, training batch sizes, weights of gradient penalties, numbers of updates, training data, expansion factors $\delta_1$, shrinkage factors $\delta_2$ and interval steps $\Delta$ are the same for all the methods involved and are listed in Table \ref{tabel:parameter}.
In G-GANs, G-GANs$^\dagger$ and G-GANs$^\ddagger$, $\lambda_1^{(0)}$ is chosen from the range [0.002, 0.004] with an increment of 0.0005, and $\lambda_2^{(0)}$ is varied within [0.01, 0.03] with a step size of 0.005 and
$\lambda_3^{(0)}$ is selected from the set $\{10^{-8}, 10^{-7}, 10^{-6}, 10^{-5}, 10^{-4}\}$.

We evaluate the performance of G-GANs in (M1)-(M4) via the MMD with
$N=M=2,000$.
We denote the estimated input dimension by Dim. and
the proportion of zero elements in the network model
by Prop.0, which is calculated by $\text{card}(i:\hat{\theta}_i=0)/\text{card}(\hat{\theta})$, where $\text{card}(\mathcal{A})$ is the cardinality of a set $\mathcal{A}$.
We report the mean and standard deviation (SD) of all indices derived from 10 experimental runs in Table \ref{result_num}. It is clear that the proposed G-GANs significantly outperforms the baseline models in terms of the MMD across all four models. Furthermore, the estimated input dimension of G-GANs is approximately 10, aligning closely with the true dimension and confirming the consistency of the input dimension, as demonstrated in Theorem \ref{t2}.
By comparing G-GAN$^{W \dagger, \ddagger}$ and G-GAN$^{SN \dagger, \ddagger}$ with WGAN-GP and SNGAN, we see that the sparse penalty on  the generator network  does not yield substantial improvement.
However, the penalization of the rows $\bB$ is effective at identifying the input dimension, although the improvements in MMD are not significant for G-GAN$^{W\ddagger}$ or G-GAN$^{SN\ddagger}$ compared to their baseline models.
However, applying additional penalties to the network depth induces notably significant benefits across all indices.
Specifically, upon examining the proportion of zero elements in network parameters for G-GANs and G-GANs$^\ddagger$, reducing the depth drastically shrinks the network size since our initial generator network is wide.
As a result, the MMD is reduced to approximately 1/10 to 1/2 of that of the baseline models across the four simulated datasets.
This finding validates Theorem \ref{t1}, highlighting the strong dependence of the generalization error on the network size.

\begin{table}[]
\centering
\spacingset{0.95}
\caption{The mean of maximum mean discrepancy (MMD), input dimension (Dim.) and proportion of zero elements in model parameters $\hat{\theta}$ (Prop.0) and the corresponding standard deviations (reported in parentheses) for (M1)-(M4). The reported MMD values have been scaled by a factor of 0.0001. The smallest MMDs, lowest input dimensions and highest portion of zero elements are highlighted in bold font.
}
\label{result_num}
\begin{tabular}{@{}c|ccc|ccc@{}}
\toprule
Dataset     & \multicolumn{3}{c}{(M1)}                                           & \multicolumn{3}{c}{(M2)}                                           \\ \midrule
Method      & MMD(SD)                 & Dim.(SD)                & Prop.0(SD)               & MMD(SD)                 & Dim.(SD)                & Prop.0(SD)               \\ \hline
WGAN-GP     & 0.25(0.04)          & 50.00(0.00)                  & 1.41(0.07)           & 4.20(4.40)            & 50.00(0.00)                  & 1.79(0.13)           \\
G-GAN$^{W\dagger}$  & 0.60(0.05)           & 50.00(0.00)                  & 13.75(0.37)          & 5.10(0.69)           & 50.00(0.00)                  & 19.51(1.10)           \\
G-GAN$^{W\ddagger}$ & 0.34(0.06)          & 11.80(6.10)           & 24.88(0.43)          & 4.86(1.20)           & 10.40(4.19)          & 32.35(2.32)          \\
G-GAN$^W$(prop.)      & \textbf{0.11(0.02)} & \textbf{11.80(6.10)}  & \textbf{91.06(0.10)}  & \textbf{0.25(0.05)} & \textbf{10.40(4.19)} & \textbf{93.51(0.18)} \\ \hline
SNGAN       & 1.03(0.15)          & 50.00(0.00)                  & 1.50(0.03)            & 2.77(0.35)          & 50.00(0.00)                  & 1.60(0.09)            \\
G-GAN$^{SN\dagger}$    & 1.07(0.17)          & 50.00(0.00)                  & 16.60(0.75)           & 2.68(0.34)          & 50.00(0.00)                  & 18.50(1.48)           \\
G-GAN$^{SN\ddagger}$   & 1.18(0.13)          & 11.00(4.20)             & 26.80(0.72)           & 2.67(0.46)          & 10.60(4.61)          & 24.30(4.32)           \\
G-GAN$^{SN}$(prop.)     & \textbf{0.14(0.01)} & \textbf{11.00(4.20)}    & \textbf{92.16(0.33)} & \textbf{0.55(0.04)} & \textbf{10.60(4.61)} & \textbf{93.29(0.16)} \\ \midrule
Dataset     & \multicolumn{3}{c}{(M3)}                                           & \multicolumn{3}{c}{(M4)}                                           \\ \midrule
Method      & MMD(SD)                 & Dim.(SD)                & Prop.0(SD)               & MMD(SD)                 & Dim.(SD)                & Prop.0(SD)               \\ \hline
WGAN-GP     & 3.57(1.27)          & 50.00(0.00)                  & 1.68(0.12)           & 1.58(0.44)          & 50.00(0.00)                  & 1.50(0.07)            \\
G-GAN$^{W\dagger}$  & 3.27(1.47)          & 50.00(0.00)                  & 17.37(1.31)          & 1.89(0.34)          & 50.00(0.00)                  & 15.88(0.74)          \\
G-GAN$^{W\ddagger}$ & 3.16(0.57)          & 9.80(4.80)            & 31.81(1.07)          & 1.56(0.23)          & 12.10(5.92)          & 28.64(0.74)          \\
G-GAN$^W$(prop.)      & \textbf{0.38(0.07)} & \textbf{9.80(4.80)}   & \textbf{90.63(0.23)} & \textbf{0.71(0.05)} & \textbf{12.10(5.92)} & \textbf{88.76(1.28)} \\ \hline
SNGAN       & 3.01(1.48)          & 50.00(0.00)                  & 1.55(0.07)           & 2.01(0.75)          & 50.00(0.00)                  & 1.54(0.08)           \\
G-GAN$^{SN\dagger}$    & 2.87(1.12)          & 50.00(0.00)                  & 16.90(0.84)           & 1.83(0.68)          & 50.00(0.00)                  & 15.80(0.90)            \\
G-GAN$^{SN\ddagger}$   & 2.74(0.64)          & 11.40(5.22)          & 27.50(1.34)           & 1.75(0.37)          & 8.30(2.93)           & 26.10(2.31)           \\
G-GAN$^{SN}$(prop.)     & \textbf{0.25(0.01)} & \textbf{11.40(5.22)} & \textbf{91.99(0.83)} & \textbf{0.49(0.02)} & \textbf{8.30(2.93)}  & \textbf{92.1(0.81)}  \\ \bottomrule
\end{tabular}
\end{table}

\subsection{The CT slice dataset}
The CT slice dataset is available from the UCI machine learning repository and comprises a collection of 53,500 CT images obtained from 74 different patients, encompassing 43 males and 31 females. Each CT slice is described by two histograms in polar space,
totaling 384 features. The first histogram delineates the location of bone structures in the image, with 240 features. The second histogram depicts the location of air inclusions within the body, with 144 features. The bins outside the image are marked with a value of $-$0.25. We partition the CT slice dataset into a training set comprising 50,000 samples and a testing set of 3,500 samples.

To assess G-GANs across diverse input dimensions and generator architectures, we applied G-GAN$^W$, G-GAN$^{SN}$, WGAN-GP, and SNGAN to the CT slice dataset using two setups: one with an initial input dimension of 64 and a 4-layer generator (each layer having a width of 256) and another with an initial input dimension of 96 and a 5-layer generator (each layer featuring a width of 320). For ease of reference, these setups are denoted as $64-4 \times 256$ and $96-5 \times 320$, respectively.
The discriminator architecture is the same across both settings, comprising 3 layers, each with a width of 192.

Common parameters persist across all methods and are outlined in Table 1.
When implementing G-GAN$^W$ and G-GAN$^{SN}$, $\lambda^{(0)}_1$ is chosen from [0.01, 0.05] in increments of 0.005,
$\lambda^{(0)}_2$ from [0.1, 0.3] in increments of 0.1 and $\lambda^{(0)}_3$ is selected in $\{10^{-8}, 10^{-7}, 10^{-6}, 10^{-5}, 10^{-4}\}$.

We calculate the MMD based on $N=M=3500$ generated and test samples.
Table \ref{result_ctslice} presents the mean and standard deviation (SD) of the MMD, input dimension (Dim.), and the proportion of zero elements (Prop.0) based on 5 experiments. As shown in Table \ref{result_ctslice}, G-GAN$^W$ and G-GAN$^{SN}$ significantly outperform the baseline models WGAN-GP and SNGAN in both scenarios. In particular, G-GAN$^W$ displays enhancements of $7.8\%$ and $26.6\%$ relative to WGAN-GP, while G-GAN$^{SN}$ exhibits improvements of $74.3\%$ and $74\%$ relative to SNGAN.
In addition, interestingly, the MMD and Dim. were nearly identical for G-GAN$^W$ and G-GAN$^{SN}$ across both scenarios.
This finding suggests that the proposed G-GANs can achieve an optimal parameter set regardless of the initial input dimension and generator architecture. Moreover, both G-GAN$^W$ and G-GAN$^{SN}$ demonstrate an input dimension of approximately 5 in both setups, aligning closely with the estimated intrinsic dimension of 3.8 by \cite{facco2017estimating}.

\begin{table}[]
\centering
\spacingset{0.95}
\caption{The mean of maximum mean discrepancy (MMD), input dimension (Dim.) and proportion of zero elements in model parameters $\hat{\theta}$ (Prop.0) and the corresponding standard deviations (reported in parentheses) for CT slices dataset. The reported MMD values have been scaled by a factor of 0.0001. The architecture is identified by the initial input dimension and generator architecture, designated as $d-l\times w$, where $d$, $l$, $w$ refer to the initial input dimension, depth and width of the generator, respectively.
The smallest MMDs, lowest input dimensions and highest portion of zero elements are highlighted in bold font.
}
\label{result_ctslice}
\begin{tabular}{@{}c|ccc|ccc@{}}
\toprule
Dataset     & \multicolumn{6}{c}{CT slices}                                                     \\ \midrule
Architecture & \multicolumn{3}{c}{$64-4\times256$}              & \multicolumn{3}{c}{$96-5\times320$}              \\ \hline
Method      & MMD(SD)          & Dim.(SD)      & Prop.0(SD)      & MMD(SD)          & Dim.(SD)      & Prop.0(SD)      \\ \hline
WGAN-GP     & 8.83(0.36)   & 64.00(0.00)        & 1.90(0.04)   & 10.95(1.52)  & 96.00(0.00)        & 2.12(0.02)  \\
G-GAN$^W$      & \textbf{8.14(0.04)}   & \textbf{5.00(1.63)}   & \textbf{59.30(0.07)}  & \textbf{8.04(0.02)}   & \textbf{5.40(1.02)} & \textbf{73.14(0.04)} \\ \hline
SNGAN       & 26.70(10.90) & 64.00(0.00)        & 2.25(0.17)  & 30.65(11.09) & 96.00(0.00)        & 2.24(0.04)  \\
G-GAN$^{SN}$     & \textbf{7.95(0.01)}   & \textbf{5.30(2.49)} & \textbf{57.99(0.35)} & \textbf{7.98(0.01)}   & \textbf{4.60(0.50)}  & \textbf{73.15(0.03)} \\ \bottomrule
\end{tabular}
\end{table}

\subsection{MNIST and FashionMNIST}

\begin{figure}
\centering
    \subfigure[Real data]{
        \begin{minipage}[t]{0.18\linewidth}
        \centering
        \label{real_mnist}
        \includegraphics[width=1.25in]{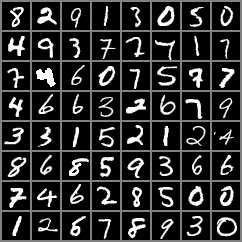}
        \end{minipage}%
    }
    \subfigure[WGAN-GP]{
        \begin{minipage}[t]{0.18\linewidth}
        \centering
        \label{wgan_gp_mnist}
        \includegraphics[width=1.25in]{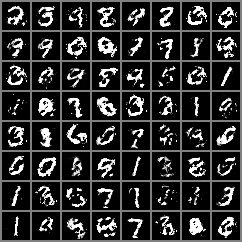}
        \end{minipage}%
    }
    \subfigure[G-GAN$^W$]{
        \begin{minipage}[t]{0.18\linewidth}
        \centering
        \label{ggan_w_mnist}
        \includegraphics[width=1.25in]{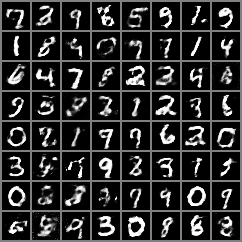}
        \end{minipage}%
    }
    \subfigure[SNGAN]{
        \begin{minipage}[t]{0.18\linewidth}
        \centering
        \label{sngan_mnist}
        \includegraphics[width=1.25in]{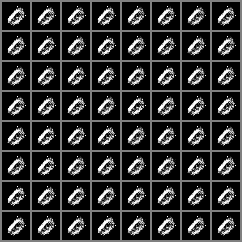}
        \end{minipage}%
    }
    \subfigure[G-GAN$^{SN}$]{
        \begin{minipage}[t]{0.18\linewidth}
        \centering
        \label{ggan_sn_mnist}
        \includegraphics[width=1.25in]{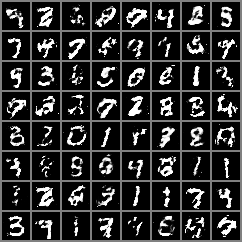}
        \end{minipage}%
    }
    \spacingset{0.95}
\caption{Observed images (a) and generated images (b) -- (e) by WGAN-GP, G-GAN$^W$, SNGAN and G-GAN$^{SN}$, respectively for MNIST. }
\label{figure:mnist}
\end{figure}

\begin{figure}
\centering
    \subfigure[Real data]{
        \begin{minipage}[t]{0.18\linewidth}
        \centering
        \label{real_fmnist}
        \includegraphics[width=1.25in]{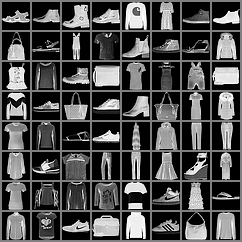}
        \end{minipage}%
    }
    \subfigure[WGAN-GP]{
        \begin{minipage}[t]{0.18\linewidth}
        \centering
        \label{wgan_gp_fmnist}
        \includegraphics[width=1.25in]{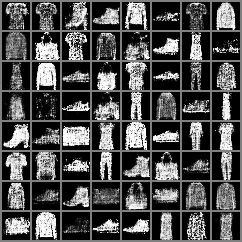}
        \end{minipage}%
    }
    \subfigure[G-GAN$^W$]{
        \begin{minipage}[t]{0.18\linewidth}
        \centering
        \label{ggan_w_fmnist}
        \includegraphics[width=1.25in]{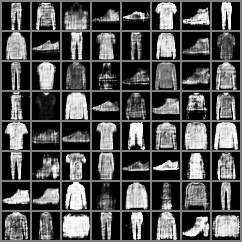}
        \end{minipage}%
    }
    \subfigure[SNGAN]{
        \begin{minipage}[t]{0.18\linewidth}
        \centering
        \label{sngan_fmnist}
        \includegraphics[width=1.25in]{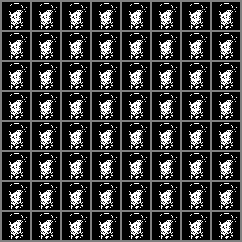}
        \end{minipage}%
    }
    \subfigure[G-GAN$^{SN}$]{
        \begin{minipage}[t]{0.18\linewidth}
        \centering
        \label{ggan_sn_fmnist}
        \includegraphics[width=1.25in]{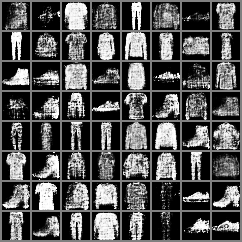}
        \end{minipage}%
    }
    \spacingset{0.95}
\caption{Observed images (a) and generated images (b)--(e) by WGAN-GP, G-GAN$^W$, SNGAN and G-GAN$^{SN}$, respectively, for FashionMNIST. }
\label{figure:fmnist}
\end{figure}

Now, we illustrate the application of G-GANs for unsupervised image generation by using two well-known benchmark datasets: MNIST and FashionMNIST. The MNIST dataset includes grayscale images of handwritten digits from 0 to 9.
FashionMNIST, an alternative to MNIST, presents a more challenging task with grayscale images
displaying ten distinct classes, such as T-shirts, dresses, shoes, and bags.
The MNIST and FashionMNIST training datasets consist of 60,000 images, each stored in a $28 \times 28$ pixel matrix with pixel intensities ranging from 0 to 1.
For visual reference, we display real examples of images from MNIST and FashionMNIST in Figure \ref{real_mnist} and Figure \ref{real_fmnist}, respectively.

We applied the proposed G-GAN$^W$ and G-GAN$^{SN}$ and their baseline models,  WGAN-GP and SNGAN, to the two datasets.
To satisfy the input requirement of the FNNs, we flatten each $28 \times 28$ image pixel matrix into a 784-dimensional vector.
The parameter settings remain consistent across all methods and are listed in Table \ref{tabel:parameter}.
When implementing G-GANs, $\lambda_1^{(0)}$ is chosen from [0.001, 0.01] in increments of 0.001 for G-GAN$^{W}$, while [0.0001, 0.0005] in increments of 0.001 for G-GAN$^{SN}$,
$\lambda_2^{(0)}$ ranges from [0.01, 0.03] in steps of 0.01 and
$\lambda_3^{(0)}$ is selected in $\{10^{-8}, 10^{-7}, 10^{-6}, 10^{-5}, 10^{-4}\}$ for both methods.

The performances of the generated samples are evaluated via FID, which is calculated based on $N=M=60,000$ generated and training samples.
Table \ref{result_f_mnist} reports the mean and standard deviation of the FID, the input dimension (Dim.) and the proportion of zero elements in the model parameters (Prop.0) by repeating the experiments 3 times.
Clearly, both G-GAN$^W$ and G-GAN$^{SN}$ outperform their baseline models, WGAN-GP and SNGAN, in terms of FID. Specifically, G-GAN$^W$ displays enhancements of $30.7\%$ in MNIST and $30.6\%$ in FashionMNIST relative to WGAN-GP; G-GAN$^{SN}$ demonstrates remarkable improvements of $55.8\%$ in MNIST and $63.2\%$ in FashionMNIST relative to SNGAN.
Furthermore, the estimated input dimension for both MNIST and FashionMNIST is approximately 8, which is consistent with the findings of \cite{pope2021intrinsic}, who estimated the intrinsic dimensions of various datasets using maximum likelihood estimation \citep{levina2004maximum}. This consistency again implies that our G-GANs are able to identify the OID. Finally, by observing a higher proportion of zero elements in G-GANs, we see that the proposed G-GANs feature simpler generator networks.

\begin{table}[]
\centering
\spacingset{0.95}
\caption{The mean of Fréchet inception distance (FID), input dimension (Dim.) and proportion of zero elements in model parameters $\hat{\theta}$ (Prop.0) and the corresponding standard deviations (reported in parentheses) for MNIST and FashionMNIST. The smallest FIDs, lowest input dimensions and highest portion of zero elements are highlighted in bold font.
}
\label{result_f_mnist}
\begin{tabular}{@{}c|ccc|ccc@{}}
\toprule
Dataset & \multicolumn{3}{c}{MNIST}               & \multicolumn{3}{c}{FashionMNIST}        \\ \midrule
Method  & FID(SD)           & Dim.(SD)      & Prop.0(SD)  & FID(SD)          & Dim.(SD)      & Prop.0(SD)   \\ \hline
WGAN-GP & 144.10(51.36) & 64.00(0.00)        & 1.31(0.27)           & 113.50(3.51)  & 64.00(0.00)        & 1.24(0.02)            \\
G-GAN$^W$  & \textbf{99.86(3.94)}   & \textbf{7.70(0.47)}  & \textbf{32.62(0.03)} & \textbf{78.65(2.01)}  & \textbf{7.7(0.47)} & \textbf{33.03(0.03)} \\ \hline
SNGAN   & 239.26(51.14) & 64.00(0.00)        & 3.19(1.06)           & 340.00(12.77)   & 64.00(0.00)        & 3.10(0.05)            \\
G-GAN$^{SN}$ & \textbf{105.70(1.45)}   & \textbf{7.70(0.47)} & \textbf{31.97(0.14)} & \textbf{125.19(2.93)} & \textbf{7.70(2.36)} & \textbf{33.23(0.08)}  \\ \bottomrule
\end{tabular}
\end{table}

To provide a more intuitive view, we show the real images and the images generated by WGAN-GP, G-GAN$^W$, SNGAN and G-GAN$^{SN}$ for MNIST and FashionMNIST in Figure \ref{figure:mnist} and \ref{figure:fmnist}, respectively. It is evident that the images generated by G-GAN$^W$ and G-GAN$^{SN}$ are much clearer than those generated by WGAN-GP and SNGAN. Particularly, when the SNGAN loses effectiveness, G-GAN$^{SN}$ still works, as shown in (d) and (e) of Figures \ref{figure:mnist} and \ref{figure:fmnist}.

\begin{figure}
\centering
    \subfigure[Thickness]{
        \begin{minipage}[t]{0.22\linewidth}
        \centering
        \label{mnist_1}
        \includegraphics[width=1.25in]{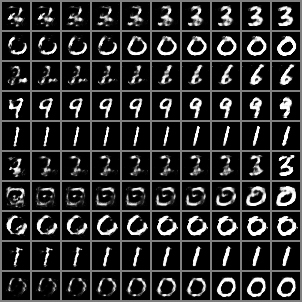}
        \end{minipage}%
    }
    \subfigure[Angle of inclination]{
        \begin{minipage}[t]{0.22\linewidth}
        \centering
        \label{mnist_2}
        \includegraphics[width=1.25in]{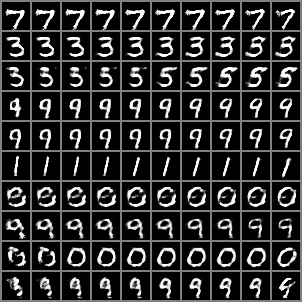}
        \end{minipage}%
    }
    \subfigure[Fabric quantity]{
        \begin{minipage}[t]{0.22\linewidth}
        \centering
        \label{fashion_1}
        \includegraphics[width=1.25in]{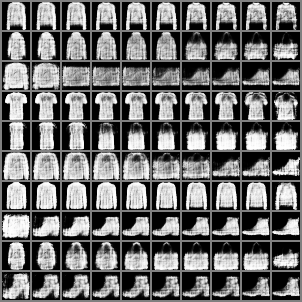}
        \end{minipage}%
    }
    \subfigure[Clothing style]{
        \begin{minipage}[t]{0.22\linewidth}
        \centering
        \label{fashion_2}
        \includegraphics[width=1.25in]{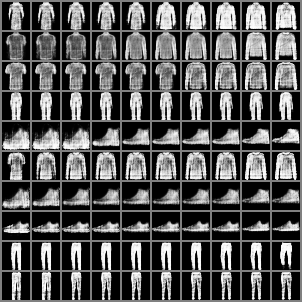}
        \end{minipage}%
    }
        \spacingset{0.95}
\caption{The manipulation of input variables in the MNIST and FashionMNIST datasets. Each block in the figures corresponds to the traversal of a single input variable while keeping the others fixed. Each row in the figures represents a different image. The traversal is conducted within the range of [-2, 2]. In MNIST, the selected input variables correspond to the thickness of the digit (a) and the angle of inclination of the digit (b). While in FashionMNIST, the selected input variables correspond to the fabric quantity (c) and the clothing style (d).}
\label{figure:interpretaion}
\end{figure}

Now, we explore the features of the input $\hat{\mathbf{B}}z$. In Figure \ref{figure:interpretaion}, we present the variation in the generated samples in MNIST and FashionMNIST as two components of the $\hat{\mathbf{B}}z$ change.
Each subfigure in Figure \ref{figure:interpretaion} corresponds to the variation in a specific input variable while holding the others constant.
Each row in the subfigures represents a different generation sample derived from the two components in $\hat{\mathbf{B}}z$ spanning from $-2$ to $2$.
Interestingly, the two chosen input variables in MNIST corresponded to the thickness (Figure \ref{mnist_1}) and the angle of inclination (Figure \ref{mnist_2}) of the digits. In FashionMNIST, the two selected input variables correspond to the fabric quantity (Figure \ref{fashion_1}) and the clothing style (Figure \ref{fashion_2}). These results indicate that the inputs identified by $\hat{\mathbf{B}}z$ align with visually prominent features, implying that when compressed to the OID, the inputs might be interpretable.
Additional experimental results regarding the interpretability of G-GAN-identified inputs for the RING, Swiss Roll, and GRID datasets \citep{chen2022inferential} can be found in the Supplementary Materials E.3.

\section{Conclusion}
\label{sec:conc}

In this study, we investigate how the input dimension impacts the generalization error of GANs.
In our theoretical analysis., we first explore the trade-off between the generator approximation and the statistical errors, confirming the existence of an OID that minimizes the generalization error of the GANs.
Moreover, the theoretical results imply that this reduction in dimensionality also minimizes the required size of the generator network architecture.
This leads to further enhancements in the accuracy and stability of the estimation and prediction results.
To determine the OID, we creatively introduce a matrix $\mathbf{B}$ into the GANs, resulting in G-GANs.
By applying a group sparse penalty to $\mathbf{B}$ and by developing an architecture penalty for the generator $g$, we establish an adaptable estimation for the input dimension and the corresponding generator architecture. Rigorous theoretical evidence supports the consistency of the proposed method in terms of both the input dimension and generator architecture.
Extensive numerical experiments, including on three benchmark datasets, demonstrate the superior performance of the proposed G-GANs in identifying the input dimension and generator architecture.
leading to a substantial improvement in both accuracy and stability for estimation and generation.
In addition, interestingly, the identified inputs align with visually significant features, such as the thickness and angle of inclination of digits in MNIST and the fabric quantity and clothing styles in FashionMNIST.

Several important research directions warrant further investigation.
First, simulation studies and three benchmark datasets suggest that the optimal input dimension is close to the intrinsic dimension $d^*$. Providing theoretical support for this observation is of interest.
Second, as illustrated in Figure \ref{figure:interpretaion} and supplementary Figure E3, when compressed to the OID, the variables $\mathbf{B}z$ seem to align with visually significant features. It is highly meaningful to further explore the structure of $\mathbf{B}$ to ensure the interpretability of $\mathbf{B}z$. Third, contrary to the notion that larger generator architectures are always preferable, our theoretical and practical analysis reveals that the ``bigger is better" belief does not hold true, which is consistent with the lottery ticket hypothesis \citep{chen2021gans, yeo2022can}.
However, there is no theoretical guarantee for this hypothesis. Establishing theoretical evidence for the lottery ticket hypothesis through generalized error decomposition is worthwhile.

\spacingset{1.1}

\bibliographystyle{apalike}
\normalem
\bibliography{Bibliography-MM-MC-mainbody}

\end{document}